\documentclass[]{dataflow}

\usepackage[toc,page,header]{appendix}

\usepackage{minitoc}
\usepackage{multirow}     
\usepackage{multicol}     
\usepackage{booktabs}     
\usepackage{colortbl}     
\usepackage[table]{xcolor} 
\usepackage{makecell}     
\usepackage{graphicx}     
\usepackage{minted}
\usepackage{xcolor}
\usepackage{fontawesome5}

\title{Towards Next-Generation LLM Training: From the Data-Centric Perspective}

\author[1]{Hao Liang}
\author[1]{Zhengyang Zhao}
\author[1]{Zhaoyang Han}
\author[1]{Meiyi Qiang}
\author[5]{Xiaochen Ma}
\author[1]{Bohan Zeng}
\author[1]{Qifeng Cai}
\author[2,4]{Zhiyu Li}
\author[2,3]{Linpeng Tang}
\author[1,2,3]{Weinan E}
\author[\ddagger,1,2,3]{Wentao Zhang}

\affiliation[]{
$^{1}$Peking University, Beijing, China \\
$^{2}$Institute for Advanced Algorithms Research, Shanghai, China \\
$^{3}$OriginHub Technology, Shanghai, China \\
$^{4}$MemTensor Technology, Shanghai, China \\
$^{5}$Hong Kong University of Science and Technology, Hong Kong SAR, China \\
}

\abstract{
Large language models (LLMs) have demonstrated remarkable performance across a wide range of tasks and domains, with data playing a central role in enabling these advances. Despite this success, the preparation and effective utilization of the massive datasets required for LLM training remain major bottlenecks. In current practice, LLM training data is often constructed using ad hoc scripts, there is still a lack of mature, agent-based data preparation systems that can automatically construct robust, reusable data workflows, thereby freeing data scientists from repetitive and error-prone engineering efforts. Moreover, once collected, datasets are typically consumed in their entirety during training, without systematic mechanisms for data selection, mixture optimization, or reweighting.
To address these limitations, we advocate two complementary research directions. First, we propose building a robust, agent-based automatic data preparation system that supports automated workflow construction and scalable data management. Second, we argue for a unified data–model interaction training system in which data is dynamically selected, mixed, and reweighted throughout the training process, enabling more efficient, adaptive, and performance-aware data utilization. Finally, we discuss the remaining challenges and outline promising directions for future research and system development.
}

\date{\today}

\def\emailicon{\raisebox{-1.5pt}{\includegraphics[height=1.05em]{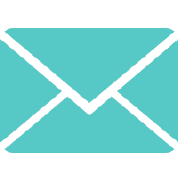}}}

\checkdata[ \emailicon \hspace{0.3em} Correspondence ]{\email{wentao.zhang@pku.edu.cn}}

\begin{document}
\maketitle

\renewcommand{\thefootnote}{\fnsymbol{footnote}} 
\setcounter{footnote}{0}

\renewcommand{\thefootnote}{\arabic{footnote}}
\pagestyle{fancy}
\fancyhf{}
\fancyhead[L]{OpenDCAI Technical Report}
\fancyhead[R]{\thepage}

\newpage
\tableofcontents
\newpage

\section{Introduction}

LLMs have exhibited exceptional performance across a wide range of tasks and domains~\cite{llama,qwen}. However, the rapid development of LLMs has significantly increased the size of training datasets~\cite{yang2024qwen2, wang2025scaling}, thereby creating an urgent need for large-scale, high-quality data. This requires large-scale data preparation and training systems~\cite{fernandez2023large, miao2024demystifying, chen2023lingua, prakash2024integrating, wang2023data}.

Although several data preparation systems—such as Data Juicer~\cite{chen2024datajuicer} and NeMo Curator—have emerged, the field still lacks a unified, user-friendly interface and agent-driven automation to streamline the entire process. Moreover, once the data is prepared, it is typically fed into the LLM training pipeline in its entirety and often in a random order, without accounting for data quality, relevance, or dynamic interaction between data and model. This static and inefficient paradigm overlooks the potential advantages of continuous data–model interaction during training. At the same time, there remains an absence of training systems capable of supporting such adaptive and feedback-driven data utilization.

In this paper, we advocate a data-centric perspective for next-generation LLM training, emphasizing two complementary principles.

\textbf{Automatic Data Preparation System.}
We argue for the necessity of an automated and reusable data preparation system to replace ad hoc, script-based workflows. To this end, we propose a layered framework consisting of an agent layer, a data operator layer, and a data serving layer. At the top level, a data agent orchestrates the end-to-end data preparation process, including operator generation, workflow composition, and standardized operator invocation. This design enables scalable, consistent, and interpretable data preparation across tasks and modalities. The backend serving layer further supports large-scale data processing, allowing the system to scale from local prototyping to distributed execution.

\textbf{Unified Data–Model Interaction Training System.}
Beyond offline data preparation, we advocate a unified data–model interaction paradigm in which data actively interacts with LLMs throughout training. In this paradigm, data selection, mixture, and reweighting are performed online and adaptively, conditioned on the model’s training dynamics. Rather than treating training data as a static resource, this approach enables model-aware data serving and more efficient utilization of large-scale datasets. We argue that such interaction can be realized through a unified training interface that accommodates diverse data selection, mixture, and reweighting strategies within a single framework.

The core contributions of this paper are summarized as follows:
\begin{itemize}
\item \textbf{New Perspective.}
We propose a novel paradigm—Next-Generation LLM Training from a Data-Centric Perspective. Specifically, we introduce an agent-guided automatic data preparation system that redefines the end-to-end workflow of LLM data preparation, enabling automation, modularity, and scalability. Furthermore, we present the vision of building Unified Data–Model Interaction Training System, where data dynamically engages with the model during training, leading to more adaptive, effective, and efficient learning processes.

\item \textbf{New Challenges and Future Directions.}  
We outline several open challenges and future research directions, including the development of domain-specific data preparation tools, the design of standardized interfaces for agent integration, the creation of efficient and intelligent data agents, and the construction of Unified Data–Model interaction systems coupled with high-performance training algorithms.  
\end{itemize}

\section{Related Works}
\subsection{Data in LLM Development}
The development of LLMs involves several key stages, among which training is particularly crucial, as the model learns fundamental linguistic patterns from large-scale corpora. During this stage, the model is exposed to vast amounts of text data from various domains, enabling it to acquire a broad understanding of language. 
Consequently, the quality and diversity of training data directly impact the model’s ability to generalize effectively across different contexts. Recently, the rapid development of large language models has brought about a substantial increase in the volume of training data~\cite{llama}. In this scenario, the quality and quantity of data become even more paramount. 

High-quality data can significantly enhance the performance of models~\cite{llama}. As the volume of data increases, ensuring high data quality becomes more challenging because it requires more resources for data cleaning, selection and annotation~\cite{bai2024survey, JCST-2509-15948}. Poor quality data can lead to models learning incorrect patterns and making inaccurate predictions. Furthermore, insufficient data diversity may result in models that perform well in specific domains but exhibit poor generalization capabilities in cross-domain tasks. Additionally, distributional shifts in data can exacerbate model reliance on training data, diminishing its applicability in real-world scenarios. 
\subsection{Data Preparation for LLMs}
As disclosed by the above discussion, data preparation is a crucial step in training LLMs, significantly impacting the model's performance and generalization capabilities. With the continuous expansion of LLM scales, the complexity and efficiency of data preparation have become key research focuses. Industrial software-based solutions like Apache Spark~\cite{Spark}, Dask~\cite{rocklin2015dask}, and Hadoop~\cite{ghemawat2003google, dean2008mapreduce, white2012hadoop} offer significant advantages in handling large-scale data, providing robust data processing and transformation capabilities. However, these solutions were not specifically designed for LLMs and may exhibit lower efficiency in certain text cleaning and labeling tasks.

\begin{figure*}
    \centering 
    \includegraphics[width=1.0\textwidth]{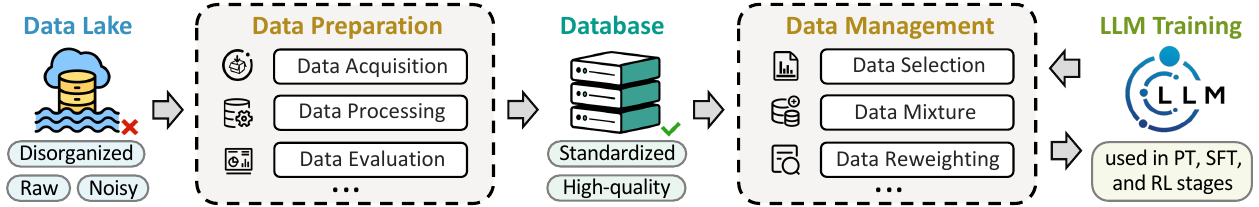}
    \caption{Illustration of Next-Generation LLM Training: The Data-Centric Paradigm.}
    \label{Fig.overview}
\end{figure*}

LLM-based methods have been widely used in data quality evaluation and data selection. For instance, MoDS~\cite{du2023mods} leverages DeBERTa for scoring and retaining high-quality data, while AlpaGasus~\cite{chen2023alpagasus} uses ChatGPT to score data accuracy. Other studies have employed GPT-4 for data rewriting and quality improvement. For a comprehensive overview, refer to the data for LLM survey~\cite{bai2024survey, JCST-2509-15948}.

Furthermore, integrated data preparation frameworks designed for LLMs significantly enhance data processing efficiency and model performance through advanced techniques. 
For instance, Data Juicer~\cite{chen2024datajuicer} and NeMo Curator are both advanced data processing frameworks for enhancing data quality in large-scale machine learning.
Data Juicer emphasizes flexibility for multilingual data preparation, while NeMo Curator, with GPU acceleration, excels in multi-modal processing and synthetic data generation, handling datasets over 100 PB. 

These solutions streamline the data processing workflow and substantially improve the generation of high-quality training data. However, they rely exclusively on filesystem-based storage without leveraging databases, which limits query efficiency. Moreover, they lack a unified interface for Data Agents to manage and operate data workflows through natural-language instructions.

\subsection{Data–Model Interaction Algorithms}\label{sec:Data-Model Interaction Algorithms}
\textbf{Online Data Selection}
Recent research has increasingly focused on data selection for large language model (LLM) training, aiming to improve efficiency and generalization by prioritizing high-value samples. Early approaches such as LESS~\cite{xia2024less} estimate each example’s influence on target objectives through gradient approximation, enabling targeted instruction tuning with only a fraction of the full dataset. LearnAlign~\cite{li2025learnalign} extends this idea to reinforcement learning settings, aligning data selection with policy-gradient directions to better capture reasoning and alignment signals. NICE~\cite{wangnice} further addresses non-differentiable metrics through black-box optimization, allowing data selection guided by discrete evaluation measures. Meanwhile, theoretical works like Data Selection via Optimal Control~\cite{gu2024data} and Data Efficacy for Language Model Training~\cite{dai2025data} formalize selection as an optimization or scoring process that quantifies data contribution and ordering effects throughout training.

\textbf{Online Data Mixture}
Recent studies have explored how to dynamically adjust domain proportions during large language model (LLM) training to improve efficiency and convergence. Aioli~\cite{chen2024aioli} models inter-domain interactions by estimating the mutual influence between domains on validation loss, allowing the mixture ratio to be updated adaptively during training. Sheared LLaMA~\cite{xia2023sheared} leverages reference losses from the original model to adjust per-domain weights, guiding training toward balanced performance after structured pruning. Efficient Online Data Mixing~\cite{albalak2023efficient} formulates data mixture as a multi-armed bandit problem, updating domain weights based on online loss feedback. Similarly, Adaptive Data Optimization~\cite{jiang2024adaptive} fits per-domain loss curves and adjusts sampling ratios according to their learning dynamics, emphasizing domains with higher marginal improvement.

In contrast, REGMIX~\cite{liu2024regmix} and DoReMi~\cite{xie2023doremi} adopt proxy-model–based strategies, estimating optimal data mixtures through regression or reference matching before large-scale training. While these static methods offer practical initialization, they lack real-time adaptability. 

\textbf{Online Data Reweighting}
Online data reweighting dynamically adjusts the importance of each training sample based on its current loss. Samples that the model finds difficult or informative are assigned higher weights, while easy or redundant ones are down-weighted as training progresses. Recent work~\cite{sow2025dynamic} demonstrates that this loss-based strategy can improve convergence speed and overall performance during large-scale pretraining. In practice, the loss serves as a simple yet effective signal for estimating data utility, which can be periodically updated during training and combined with other feedback signals—such as gradient or alignment information—within a unified data–model interaction framework.

Although many data–model interaction algorithms have been proposed in the machine learning community, they generally lack a unified, data-centric framework for systematic management and integration. To address this gap, we propose the concept of a Data–Model Interaction System, which provides an organized and scalable foundation for coordinating these methods within a consistent data-centric paradigm.

\section{Data-Centric LLM Training}
Our proposed position—Next-Generation LLM Training: From the Data-Centric Perspective integrates two stages to automatically prepare data from the data lake and use it for LLM training. As shown in Figure~\ref{Fig.overview}, the process begins with the Data Lake, which stores raw, unorganized, and noisy data. This is followed by an Automatic Data Preparation Framework, where data collection, processing, and evaluation are performed to standardize and improve data quality. The prepared data is then stored in the Database, ensuring it is high-quality and well-organized for easy access. Then, during model training, the model can interact with data for optimal data serving, including optimized data selection, mixture, and reweighting during model training. We will further introduce the automatic data preparation framework in Section~\ref{sec: Preparation_Framework}. Then we will introduce the unified data–model interaction training system in Section~\ref{sec:Data-Management}.

\begin{figure*}
    \centering
    \includegraphics[width=1.0\linewidth]{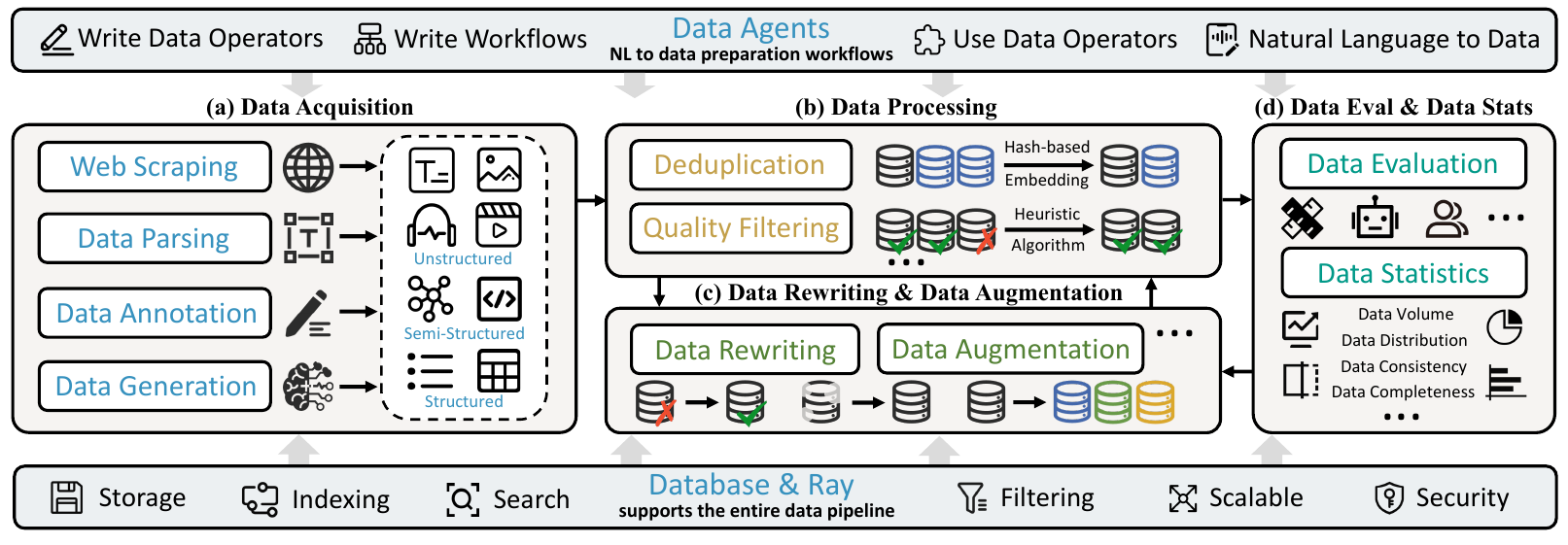}
    \caption{The Illustration of the Automatic Data Preparation System.}
    \label{Fig.data_preparation}
\end{figure*}
\section{Automatic Data Preparation System}\label{sec: Preparation_Framework}
Data preparation plays a critical role in the success of LLMs~\cite{bai2024survey}. In this section, we present a vision of a comprehensive data preparation framework organized into three hierarchical layers. At the bottom layer, the data serving layer is responsible for large-scale data storage, indexing, and efficient processing. The middle layer comprises a diverse set of data operators that perform various data preparation tasks, such as filtering, transformation, augmentation, and quality assessment. At the top layer, a Data Agent orchestrates these operators through natural language instructions, enabling flexible, automated, and human-interpretable control over the entire data preparation pipeline. For better clarity and ease of understanding, we first introduce the data operators in detail. We then describe the design and functionality of the data serving layer. Finally, we discuss how data agents can be designed to effectively coordinate data operators and support scalable, adaptive data preparation.

\subsection{Effective Data Operators}
Data Operators consist of four key phases: \textbf{Data Acquisition}, \textbf{Data Processing}, \textbf{Data Rewriting and Augmentation}, and \textbf{Data Evaluation and Statistics}, as shown in Figure \ref{Fig.data_preparation}. 

\subsubsection{Data Acquisition}
Effective data collection is essential for the training of LLMs, which require vast amounts of data. The strategies we use for data collection can be broadly categorized as follows:
 
\textbf{Web Scraping:} Data collected from the web that reflects the natural distribution of inputs encountered during model deployment is a key component. This data should be diverse and authentic, capturing a broad spectrum of scenarios to ensure the model generalizes well in real-world applications.

\textbf{Data Parsing:} Data parsing involves the extraction of information from raw data sources, such as logs, databases, or documents~\cite{wang2024mineru}, and converting it into a structured or usable format. Parsing ensures proper organization of the data for the next steps in processing.

\textbf{Data Annotation:} Data annotation is the process of labeling data with relevant tags, categories, or ground-truth values. This can be done manually by experts or through automated techniques, such as leveraging pre-trained models to speed up the process~\cite{cai2025lovr, guo2025brace}. High-quality annotations are critical for supervised learning tasks.

\textbf{Data Synthesis:} With the rise of LLMs, synthetic data has become an indispensable component. By leveraging diverse prompts~\cite{xu2024magpie, ge2024scaling} and structured templates, high-quality training datasets can be synthesized efficiently and effectively~\cite{liu2024synthvlm, sun2025mm, liang2024synth}.


\subsubsection{Data Processing}
Since data is collected from multiple sources, the next crucial step is data processing, which filters, cleans and refines the raw data for improved data quality. This data processing procedure includes multiple modules. In the following, we elaborate on two representative key modules.

\textbf{Data Deduplication:} The data deduplication process removes redundant or duplicate instances from the dataset, ensuring that the data has enough diversity. Commonly used algorithms such as minhash or embedding, and cluster-based deduplication methods~\cite{bai2024survey, JCST-2509-15948}. 


\textbf{Data Quality Filtering:} Data filtering refers to the process of removing noisy, irrelevant, or low-quality data points. Common approaches include rule-based methods and LLM-based scoring, where each data point is assigned a binary score of 0 or 1. A score of 0 indicates that the data point has been filtered out, while a score of 1 means it has been retained~\cite{shen2025let}. 



\subsubsection{Data Rewriting and Augmentation}
After data processing, the data may require further rewriting and augmentation to enhance its quality and diversity. These two steps serve distinct yet complementary roles in improving the overall effectiveness of training data.

\textbf{Data Rewriting} focuses on improving the linguistic and structural quality of the data. It involves paraphrasing sentences, refining grammar and clarity, and reformulating content to ensure coherence and consistency. Rewriting can also adjust tone or style to match task-specific requirements, reduce noise or redundancy, and correct semantic inconsistencies, ultimately aligning the data more closely with the model’s intended learning objectives.

\textbf{Data Augmentation}, on the other hand, aims to expand data diversity and robustness by generating new, meaningful variations of existing samples. This may include techniques such as synonym substitution, back-translation, multimodal transformations (e.g., caption rewriting or image-text pairing), and context expansion~\cite{cai2025text2sql}. By exposing the model to a broader range of linguistic and contextual variations, augmentation improves generalization and reduces overfitting, leading to more resilient model performance across tasks and domains.



\subsubsection{Data Quality Evaluation and Statistics}

To ensure that the dataset is of high quality and suitable for training, we employ a data quality evaluation process.

\textbf{Data Quality Evaluation.}
To guarantee the dataset is of high quality and suitable for training, we perform a comprehensive data quality evaluation that combines metric-based, LLM-based, and human-based assessments. Metric-based evaluation focuses on format correctness, duplication rate, diversity, linguistic fluency, factual accuracy, and—where applicable—educational value. LLM-based evaluation leverages a strong reference model to score samples on attributes such as coherence, helpfulness, and relevance. Human-based evaluation involves manual inspection of representative subsets to verify the effectiveness of automatic evaluations and to capture nuanced dimensions such as tone, clarity, and domain-specific appropriateness.

\textbf{Statistical Information.}
Key statistics are also provided to present an overview of the dataset’s structure and quality. These include the total number of samples, average and standard deviation of input/output lengths, distribution across different domains or tasks, proportions of retained versus filtered data, and duplication rate. Such statistics help quantify the diversity, balance, and cleanliness of the dataset, and serve as an important reference for both model training and future dataset refinement.

\begin{figure}[t]
    \centering
    \includegraphics[width=0.8\linewidth]{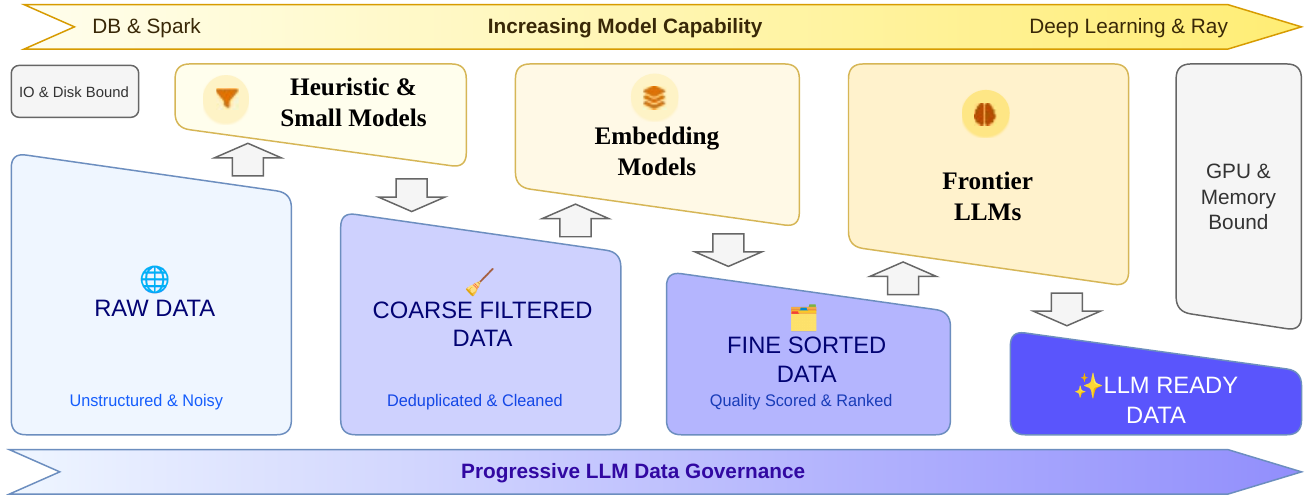}
\caption{A progressive LLM data governance pipeline exhibiting a bidirectional funnel pattern: decreasing data volume and increasing model capability, accompanied by a shift from I/O- and CPU-bound processing to GPU- and memory-bound workloads.}
\label{fig:llm_data_governance_pipeline}
\end{figure}

\subsection{Scalable Operator Serving}
Data governance operators rely on heterogeneous models to transform, synthesize, evaluate, or filter data. As shown in Fig.~\ref{fig:llm_data_governance_pipeline}, LLM data governance typically follows a bidirectional funnel pattern: as the pipeline progresses, data volume decreases while model capability increases. Accordingly, the system bottleneck shifts from disk-, CPU-, and SQL-intensive processing to deep-learning-based execution that is increasingly GPU- and memory-bound.

Not all stages are equally model-centric. In early-stage large-scale preprocessing, such as corpus cleaning, scan-heavy filtering, and dataset-wide deduplication, traditional data-processing systems such as databases and Spark~\cite{Spark} are often more effective. In contrast, later-stage operations, including semantic evaluation, data synthesis, and multimodal quality assessment, are increasingly dominated by large-model inference and accelerator efficiency.

The core systems challenge is to balance different resource bottlenecks across stages. Earlier stages process much larger data volumes with relatively lightweight computation, favoring minimal intermediate materialization and more aggressive operator fusion to reduce I/O and data movement. These operators and small models can also typically be started on demand with low overhead, and are better matched to high-throughput hardware with limited memory, such as consumer GPUs or CPU-centric clusters. In contrast, later LLM-centric stages process less data but incur much higher per-sample cost. Since large models often have expensive startup and loading overhead, the system should materialize intermediate results more aggressively and retain model workers to amortize initialization cost over as much data as possible. These stages are further constrained by model size, KV-cache footprint, and long-context execution, making large-memory accelerators such as H200 GPUs more suitable.

For the LLM-centric stages, the system should maximize reuse of existing model-serving infrastructure rather than introducing specialized execution stacks for each operator. Large models are served with optimized inference engines such as vLLM~\cite{kwon2025vllm}, vLLM-Omni, and SGLang~\cite{zheng2024sglang}, while distributed runtimes such as Ray~\cite{moritz2018ray} provide resource orchestration and parallel execution across heterogeneous hardware. In addition, accurate operator profiling and pipeline-level scheduling are critical for deciding operator fusion, task decomposition, and resource placement, thereby improving end-to-end throughput.

Overall, scalable operator serving is fundamentally a pipeline-level resource management problem. The system must jointly optimize materialization, operator fusion, hardware allocation, and scheduling according to the workload characteristics of each stage.

\subsection{Data Agent for Automatic Data Preparation}
This section outlines the vision for designing a unified Data Agent framework that seamlessly integrates with the Data Preparation System for automatic data preparation. 

\subsubsection{Data Agent Design}

Building upon the unified operator interface, we further design a \textbf{Data Agent} that enables natural-language–driven data preparation by reasoning over available system capabilities.

\textbf{Skill Abstraction.}
In our framework, both data operators and data pipelines are abstracted as reusable \emph{skills}. 
Each skill is described through structured metadata, including its functionality, input–output schema, execution constraints, and natural-language descriptions. 
This skill-oriented abstraction exposes the system capabilities to the agent in a structured yet interpretable manner, allowing the Data Agent to discover, reason about, and compose operators and pipelines dynamically.

Through this design, the agent can leverage existing operators, reuse previously defined pipelines, or extend the system by synthesizing new skills when necessary.

\textbf{Data Agent Capabilities.}
Based on the skill abstraction, the Data Agent supports several key functionalities:

\begin{enumerate}
    \item \textbf{Automatic operator generation}, where the agent synthesizes new data operators when existing capabilities are insufficient.
    \item \textbf{Workflow construction and management}, where the agent composes multiple operators into complete data processing pipelines.
    \item \textbf{Prompt composition for LLM-based operators}, enabling operators that interact with large model services to be dynamically configured through prompt generation.
    \item \textbf{Document question answering}, allowing the agent to retrieve and reason over system documentation to understand operator usage, pipeline design patterns, and execution constraints.
\end{enumerate}

These capabilities allow users to accomplish complex data preparation tasks through natural language instructions without requiring manual scripting or detailed configuration.

\textbf{Data Agent System Architecture.}
To balance reliability and flexibility in agent-driven data preparation, we envision a hybrid architecture that combines workflow-based execution with agentic reasoning.

On one hand, workflow structures provide reliability, controllability, and interpretability. 
The reasoning and execution process can be represented as a graph-structured workflow (e.g., LangGraph), where nodes correspond to reasoning or execution steps and edges encode explicit data and control dependencies. 
This structured representation ensures deterministic execution, transparent data flow, and easier debugging of complex pipelines.

On the other hand, agentic reasoning enables flexible decision-making and dynamic capability extension. 
The Data Agent can reason over available operators and pipelines, dynamically composing workflows or generating new operators when existing capabilities are insufficient.

Overall, this hybrid design integrates the robustness of workflow orchestration with the flexibility of agentic reasoning, enabling scalable, extensible, and transparent data preparation systems.

\textbf{Post-generation Human-In-the-Loop Validation.}
After the Data Agent proposes a data processing pipeline, a two-stage validation process ensures reliability and interpretability.
First, automatic validation verifies operator connectivity and data flow consistency, such as key alignment and dependency integrity.
Second, human validation allows users to assess the pipeline’s quality, task suitability, and high-level design intent.

Importantly, observations and feedback from human validation can be fed back to the Data Agent to further revise or refine the generated workflow, enabling iterative improvement rather than one-shot approval.
Beyond simple acceptance or rejection, such feedback provides corrective signals that enhance robustness, trustworthiness, and alignment with user intent.
Once validated, the system executes the workflow automatically, achieving an efficient and transparent data preparation process that balances scalable automation with human oversight.

\begin{figure*}
    \centering 
    \includegraphics[width=1.0\textwidth]{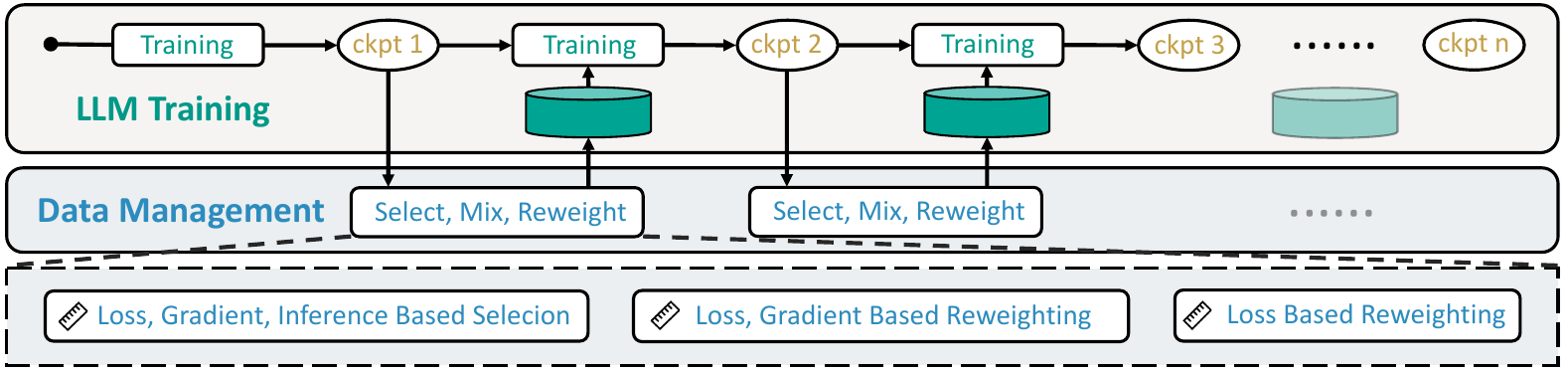}
    \caption{Illustration of the Unified Data–Model Interaction Training System.}
    \label{Fig.data_management}
\end{figure*}

\subsubsection{End-to-End Example of Agent-Driven Data Preparation}
We illustrate the proposed Data Agent with an end-to-end example that shows how natural-language instructions are translated into an executable data preparation workflow.

\textbf{User Instruction.}
A user provides a high-level request:
\emph{“Prepare a high-quality instruction-tuning dataset for mathematical reasoning. Remove low-quality and duplicate samples, augment difficult problems, and report basic dataset statistics.”}

\textbf{Agent Workflow Construction.}
In response, the Data Agent retrieves relevant operators (e.g., deduplication, quality filtering, augmentation, and statistics) via operator-aware retrieval and composes them into an ordered workflow based on operator metadata and dependency constraints:
\[
\texttt{Deduplication} \rightarrow \texttt{Quality Filtering} \rightarrow \texttt{Data Augmentation} \rightarrow \texttt{Data Statistics}.
\]
Each operator is instantiated with agent-generated prompts and parameters, and automatically validated for interface compatibility and data-flow consistency.

\textbf{Human-in-the-Loop Refinement and Execution.}
Before execution, the generated workflow is presented for human inspection. User feedback (e.g., reducing aggressive augmentation for easy samples) is fed back to the Data Agent, which revises the workflow accordingly while preserving its overall structure. After validation, the refined workflow is executed automatically, producing a cleaned and augmented dataset together with a reusable and transparent workflow specification.

\section{Unified Data–Model Interaction Training System}\label{sec:Data-Management}
After data preparation, we turn to introduce a unified training system that supports dynamic data–model interaction and describe its overall design.

\subsection{Unified Data–Model Interaction Training System Design}
The proposed data–model interaction system centers on three fundamental components: data selection, data mixture, and data reweighting. As summarized in Section~\ref{sec:Data-Model Interaction Algorithms}, we identify the shared algorithmic abstractions underlying these approaches and design unified training interfaces that enable a coherent, extensible, and model-agnostic data–model interaction framework.

\textbf{Data Selection Module.}
As illustrated in Figure~\ref{Fig.data_management}, the envisioned system supports dynamic data selection conditioned on the current model checkpoint during training. To enable algorithm-agnostic integration of data selection methods, we introduce a unified interface between the training process and the data selection module. Specifically, the training system exposes three core capabilities:
(1) access to per-sample or subset-level gradients,
(2) model inference on specified data instances, and
(3) computation of data-dependent training loss.
Based on these signals, the data selection module outputs a binary selection mask indicating whether each data instance is selected or discarded for the subsequent training step.

This abstraction captures the common requirements shared by a wide range of existing data selection algorithms. We focus on these three signals because they cover the core information used by most practical data selection methods. In contrast, higher-order information such as Hessian or quasi-second-order approximations is prohibitively expensive to compute for large language models at scale, and is therefore not considered in the current system design.

\textbf{Data Mixture Module.}
The data mixture module dynamically adjusts the sampling proportions of different domains during training. As shown in Figure~\ref{Fig.data_management}, we abstract data mixture strategies as loss- or gradient-based methods operating at the domain level.

Accordingly, the training system exposes two key capabilities:
(1) access to per-sample or subset-level gradients, and
(2) computation of data-dependent training loss.
Using these signals, the data mixture module estimates domain-level importance weights, which are then used to update the sampling distribution over domains. By periodically re-estimating inter-domain influence, the system increases the sampling probability of underrepresented or challenging domains while downsampling redundant or overrepresented data, leading to more balanced and effective training.

\textbf{Data Reweighting Module.}
The data reweighting module adjusts the contribution of individual samples during training. Each data instance is assigned a continuous weight.

Accordingly, the training system exposes a key capability to the reweighting module:
(1) computation of data-dependent training loss.
Based on this signal, the data reweighting module estimates sample-level importance weights.

These weights are continuously updated as training progresses, enabling the system to emphasize informative or hard samples while downweighting noisy, redundant, or outdated data. To support this process in a systematic and extensible manner, all reweighting methods are integrated through a unified data-weighting interface abstracted from the trainer, ensuring consistent integration of diverse reweighting strategies within the overall system.

With the data selection, data mixture, and data reweighting modules in place, the proposed Data–Model Interaction System provides unified management of computational resources and training workflows, enabling data adaptation and model optimization to be jointly scheduled, monitored, and adjusted. At the same time, the framework offers a unified experimental environment for systematically studying and comparing data selection, mixture, and reweighting strategies under consistent training settings. 




\section{Future Directions}
This section outlines several open challenges and promising directions for advancing both the \textbf{Automatic Data Preparation System} and the \textbf{Unified Data–Model Interaction Training System}.

\subsection{Automatic Data Preparation System}

\textbf{Ease of Use and Unified Framework.}
Ease of use remains a fundamental challenge in large-scale data preparation. Data scientists and engineers require intuitive, high-level abstractions that allow them to orchestrate complex data pipelines without extensive manual intervention. A unified data agent system is therefore essential—one that integrates automation, modular operators, and transparent execution, enabling agents to coordinate heterogeneous data tasks such as collection, cleaning, labeling, and evaluation within a single framework. 

\textbf{Data Agents and Agent Systems.}
A well-designed data preparation framework should adopt a hierarchical agent architecture. High-level agents are responsible for planning, scheduling, and coordinating data workflows, while low-level agents execute domain-specific operations through standardized interfaces. This agent-based abstraction can substantially reduce human workload while ensuring that data pipelines remain adaptive, interpretable, and reusable across tasks, domains, and projects.

\textbf{Advanced Data Preparation Algorithms.}
Beyond system abstractions, progress in automatic data preparation also requires advances in data processing algorithms. The community should develop cutting-edge methods that leverage statistical, linguistic, and multimodal signals to assess, transform, and synthesize data at scale. These algorithms must integrate seamlessly with agent workflows and data backends, supporting large-scale parallelism and continuous data updates. Ultimately, the goal is to build an ecosystem in which intelligent agents collaborate through standardized operator interfaces to realize fully automated, domain-aware data pipelines.

The DataFlow Ecosystem~\cite{liang2025dataflow} offers an initial exploration of this direction, illustrating how unified data preparation systems, data agents, and scalable preparation algorithms can be co-designed within a single ecosystem.

\subsection{Unified Data–Model Interaction Training System}

\textbf{Data-Centered Computational Support.}
The next frontier of LLM training demands data-centered computational primitives and infrastructure support. While existing deep learning frameworks are highly optimized for model throughput, they typically lack fine-grained access to data-level signals such as per-sample gradients, influence estimates, or backward–inference interactions. Future systems should extend large-scale training frameworks (e.g., DeepSpeed~\cite{rasley2020deepspeed} or Megatron-LM~\cite{shoeybi2019megatron}) to natively support data-level differentiation and model–data co-optimization, enabling efficient coupling between model computation and data feedback.

\textbf{Scalable and Efficient Algorithms.}
Achieving adaptive data–model interaction at scale requires algorithms that remain computationally efficient under massive data regimes. Many existing methods, while effective in small-scale or offline settings, rely on expensive operations such as repeated inference, loss evaluation, or gradient computation, which become prohibitively costly in large-scale training. In particular, a number of data selection and influence estimation approaches depend on second-order information or per-sample gradient statistics to assess the contribution of individual training examples. While these signals can provide fine-grained insights into data utility, computing and maintaining them at scale is extremely challenging, especially in trillion-token training scenarios. On the other hand, approaches that avoid such detailed information often struggle to accurately estimate data importance, resulting in suboptimal data scheduling or selection strategies. Therefore, designing scalable algorithms that balance computational efficiency with informative signals for data valuation remains a central challenge for adaptive data–model interaction.

The DataFlex training system~\cite{dataflex2026} represents an initial exploration of this direction, demonstrating how data and models can interact within a unified training framework.

\section{Conclusion}\label{sec: Conclusion}
In this paper, we presented a data-centric perspective on next-generation LLM training, highlighting the limitations of current data preparation and utilization practices. We argued that ad hoc, script-based pipelines and static, one-shot data consumption are increasingly misaligned with the scale and complexity of modern LLM training.
To address these challenges, we outlined two complementary directions: an agent-based data preparation system that enables automated, robust, and reusable data workflows, and a data–model interaction paradigm that treats data as a dynamic and adaptive resource throughout training. Together, these directions aim to shift data from a passive input to an active component in the learning process, improving both efficiency and effectiveness in large-scale model training.
We believe that advancing these ideas requires joint progress in system design, algorithmic abstraction, and human-in-the-loop control. By articulating these directions and their associated challenges, we hope to encourage further research on principled, scalable, and agent-driven data systems for large language models.

\clearpage

\bibliographystyle{plainnat}
\bibliography{main}






\end{document}